\pdfoutput=1
\documentclass[11pt]{article}

\usepackage[final]{acl}

\usepackage{times}
\usepackage{latexsym}

\usepackage[T1]{fontenc}
\usepackage[utf8]{inputenc}

\usepackage{microtype}

\usepackage{inconsolata}

\usepackage{graphicx}
\usepackage{enumitem}
\usepackage{hyperref}
\usepackage{url}
\usepackage{booktabs}
\definecolor{darkblue}{rgb}{0, 0, 0.5}
\hypersetup{colorlinks=true, citecolor=darkblue, linkcolor=darkblue, urlcolor=darkblue}

\usepackage{xspace}
\usepackage{wrapfig}
\usepackage{float}
\usepackage{natbib}
\usepackage{xcolor}
\usepackage{amsmath}
\usepackage{amsfonts}
\usepackage{algorithm}
\usepackage{algorithmicx}
\usepackage{algpseudocode}
\usepackage{spverbatim}
\usepackage{soul}
\usepackage{tablefootnote}
\usepackage{multirow}
\usepackage[textsize=scriptsize]{todonotes}
\usepackage{colortbl}
\usepackage{yfonts}
\usepackage{amssymb}
\usepackage{wasysym}
\usepackage{longtable}
\usepackage{makecell}
\usepackage{subcaption}
\usepackage{twemojis}
\usepackage{tabularx} % For tabularx environment
\usepackage{fontawesome}
\usepackage{array}    % For table wrapping
\definecolor{darkgreen}{rgb}{0.0, 0.5, 0.13} 
\definecolor{gold}{rgb}{0.83, 0.69, 0.52}
\definecolor{lightred}{HTML}{e99090}

\newcommand{\claude}{\textsc{\small Claude Sonnet 3.7}}

\newcommand{\gpt}{\textsc{\small GPT-4}}
\newcommand{\gptmini}{\textsc{\small GPT-4o-mini}}
\newcommand{\oone}{\textsc{\small o1}}
\newcommand{\othree}{\textsc{\small o3}}
\newcommand{\othreemini}{\textsc{\small o3-mini}}
\newcommand{\gpto}{\textsc{\small GPT-4o}}
\newcommand{\gptfourone}{\textsc{\small GPT-4.1}}
\newcommand{\deepseek}{\textsc{\small DeepSeek-R1}}
\newcommand{\geminiproone}{\textsc{\small Gemini Pro 1.5}}
\newcommand{\geminiprotwo}{\textsc{\small Gemini Pro 2.5}}

\newcommand{\llamasmall}{\textsc{\small Llama 3.1 Instruct (8B)}}
\newcommand{\llamalarge}{\textsc{\small Llama 3.3 Instruct (70B)}}
\newcommand{\qwensmall}{\textsc{\small Qwen 2.5 Instruct (7B)}}
\newcommand{\qwenlarge}{\textsc{\small Qwen 2.5 Instruct (72B)}}
\newcommand{\qwenthreesmall}{\textsc{\small Qwen 3 (8B)}}
\newcommand{\qwenthreelarge}{\textsc{\small Qwen 3 (32B)}}
\newcommand{\gteqwenparam}{\textsc{\small gte-Qwen2-7B-instruct (7B)}}
\newcommand{\gteqwen}{\textsc{\small gte-Qwen2-7B-instruct}}
\newcommand{\human}{\textsc{\small Human}}

\newcommand\blankfootnote[1]{%
  \let\thefootnote\relax\footnotetext{#1}%
  \let\thefootnote\svthefootnote%
}

\title{\raisebox{-0.3\height}{\includegraphics[width=1.75cm]{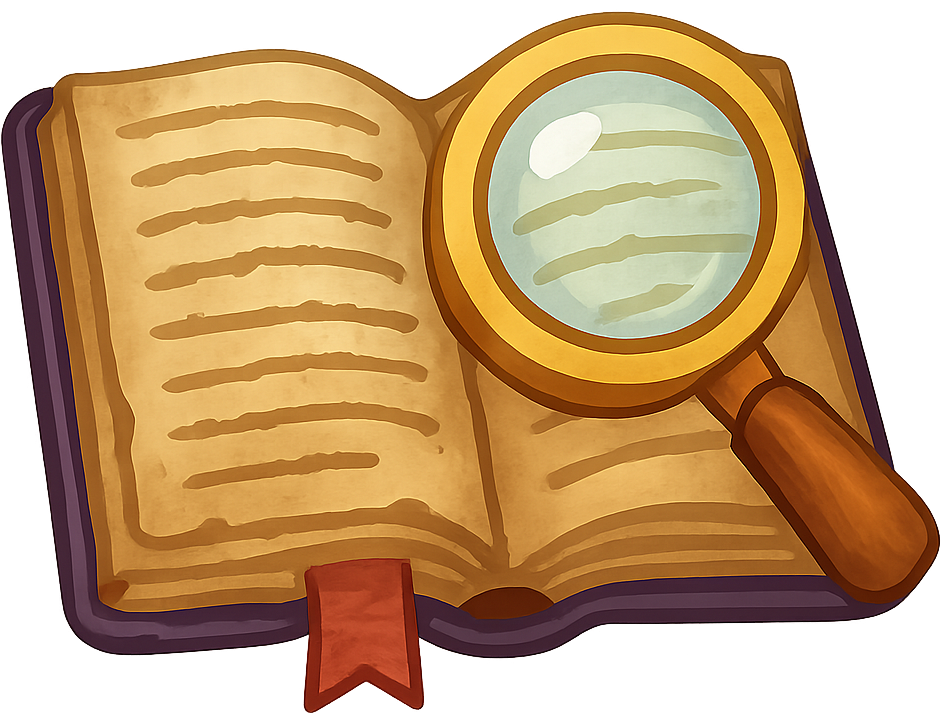}}\; Literary Evidence Retrieval via Long-Context Language Models}

\author{Katherine Thai\twemoji{crystal ball} and Mohit Iyyer\twemoji{crystal ball}\twemoji{magnifying glass tilted right}\\
UMass Amherst\twemoji{crystal ball}\hspace{0.3cm} 
University of Maryland, College Park\twemoji{magnifying glass tilted right},\\
\texttt{kbthai@umass.edu}, \texttt{miyyer@umd.edu}  \\
\\
\url{https://github.com/katherinethai/long_context_relic}
}

\begin{document}

\maketitle

\begin{abstract}

How well do modern long-context language models understand literary fiction?  We explore this question via the task of \emph{literary evidence retrieval}, repurposing the RELiC dataset of~\citet{thai-etal-2022-relic} to construct a benchmark where the entire text of a primary source (e.g., \textit{The Great Gatsby}) is provided to an LLM alongside literary criticism with a missing quotation from that work. This setting, in which the model must generate the missing quotation, mirrors the human process of literary analysis by requiring models to perform both global narrative reasoning and close textual examination. We curate a high-quality subset of 292 examples through extensive filtering and human verification. Our experiments show that recent reasoning models, such as \geminiprotwo\ can exceed human expert performance (62.5\% vs. 50\% accuracy). In contrast, the best open-weight model achieves only 29.1\% accuracy, highlighting a wide gap in interpretive reasoning between open and closed-weight models. Despite their speed and apparent accuracy, even the strongest models struggle with nuanced literary signals and overgeneration, signaling open challenges for applying LLMs to literary analysis. We release our dataset and evaluation code to encourage future work in this direction.

\end{abstract}
\section{Introduction}
\label{sec:intro} 

The emergence of long-context language models, which can process millions of tokens~\cite{geminiteam2024gemini}, has unlocked new AI applications for literary analysis. In this paper, we focus on the task of \emph{literary evidence retrieval}, in which a model must retrieve a supporting quotation from a primary source (e.g., a novel) to substantiate an excerpt of literary criticism.

~\citet{thai-etal-2022-relic} frame literary evidence retrieval as a computational task by introducing RELiC, a dataset that contains short excerpts from published literary criticism that include quotations from famous novels. While RELiC was developed to benchmark and improve retriever models, which compute embeddings of claims and short chunks of the book text, we repurpose it as a testbed for long-context\footnote{In this paper, we consider ``long-context" to mean allowing a minimum of 128k tokens as input.} language models: A model is given the entire text of the book and an excerpt of literary criticism with a missing quotation from that book, then asked to generate the missing quote (see \autoref{fig:llm_fail}, top). This task does not exactly mirror the human process of literary analysis, in which a scholar typically develops claims and selects supporting quotes iteratively. However, the task requires the same understanding of and complex reasoning over plot, subtext, and other literary devices with the advantage of being easily verifiable. 

To enable robust evaluation, we curated a high-quality subset of 292 examples from RELiC, verified through a combination of automated filtering and expert human review. These 292 examples contain claims that require both global reasoning over events and ``close reading'' over singular passages to solve. Our experiments reveal that state-of-the-art reasoning models significantly outperform previous LLMs and even surpass a human expert baseline: our best model, \geminiprotwo\ obtains an accuracy of \textbf{62.5\%} compared to \textbf{55.0\%} for a human expert on a subset of the data. However, we also find that these models tend to overgenerate and struggle with subtle literary cues. Open-weight models perform substantially worse, suggesting that interpretive reasoning, not just long-context capacity, is essential to success in this domain, forming important directions for future research.

\begin{table}[t]
\centering
\footnotesize % Reduce font size for fitting in one column

\centering

\resizebox{0.48\textwidth}{!}{%
\begin{tabular}{lcc|cc|c}
\toprule
 & \multicolumn{2}{c}{\textbf{Primary Sources }} & \multicolumn{3}{c}{\textbf{Dataset Examples}} \\
 & \multicolumn{2}{c}{(\textit{n=7})} & \multicolumn{3}{c}{(\textit{n=292})} \\
\cmidrule(lr){2-3} \cmidrule(lr){4-6}
 & \textsc{Tokens} & \textsc{Words} & \textsc{Tokens} & \textsc{Words} & \textsc{\# Ex/Book}\\
\midrule
\textsc{Mean} & 85,526 & 69,456 & 254.9 & 203.6 & 36.0 \\
\textsc{St. Dev.} & 26,6304 & 21,167 & 66.9 & 52.9 & 19.6 \\
\textsc{Max} & 124,544 & 102,549 & 492.0 & 385.0 & 52.0 \\
\textsc{Min} & 45,038 & 37,209 & 116.0 & 91.0 & 7.0 \\
\bottomrule
\end{tabular}
}
\caption{Summary statistics for long-context RELiC. Token counts were computed with the \texttt{o200k\_base} encoding via \texttt{tiktoken} (\url{https://github.com/openai/tiktoken}) and word counts were computed by splitting on whitespace.}
\label{tab:basic_summary_stats_dataset}
\end{table}

\section{Dataset Curation}
\label{sec:data_methods}

\begin{table}[t!]
\small
\resizebox{0.48\textwidth}{!}{%
\begin{tabular}{lccc} % Updated to four columns

\toprule
\textsc{Model + Simple Prompt} & \textsc{ALL} \textsubscript{($n$=292)} &  \textsc{\faUser} \textsubscript{($n$=40)}& \textsc{\faBook} \textsubscript{($n$=39)}\\ 
\midrule
\geminiprotwo~\cite{google2025gemini25} & 63.7 & 57.5 & \textbf{82.1} \\
\othree~\cite{o3} & 49.7 & 47.5 & 71.8 \\
\geminiproone~\cite{geminiteam2024gemini} & 36.1  & 22.5  & 43.6 \\
\oone~\cite{openai2024o1}  & 32.2  & 25.0  & 41.0 \\
\gpto\ \cite{openai2024gpt4ocard} & 27.4  & 17.5  & 35.9 \\
\qwenlarge\ \cite{qwen_qwen25_2025} & 11.3 & 7.5  & 20.5 \\
\llamasmall\ \cite{grattafiori2024llama3herdmodels} & 5.1  & 5.0  & 2.6 \\
\qwensmall & 2.7  & 5.0  & 2.6  \\
\llamalarge & 2.1  & 0.0  & 2.6 \\
\midrule
\textsc{Model + Explanation Prompt} & & & \\
\midrule
\geminiprotwo & \textbf{64.7} & \textbf{62.5} & 79.5 \\
\gptfourone~\cite{gpt-4.1} & 51.0 & 47.5 & 69.2 \\
\othree & 50.7 & 50.0 & 66.7 \\
\geminiproone & 38.5  & 40.0  & 50.0 \\
\claude~\cite{claude_sonnet_2025} & 37.0 & 32.5 & 48.7 \\
\deepseek~\cite{deepseekai2025deepseekv3technicalreport} & 29.1 & 15.0 & 38.5 \\
\gpto & 24.3  & 22.5  & 31.8 \\
\qwenthreelarge~\cite{qwen_qwen25_2025} & 19.2 & 20.0 & 33.3 \\
\qwenthreesmall  & 8.9 & 5.0 & 10.3 \\
\othreemini\ \cite{openai2025o3mini} & 8.3  & 10.0  & 13.6 \\
\midrule
\textsc{Baselines} & & & \\
\midrule
\gteqwen & 4.5  & 2.5 & 6.8 \\
\human & - & 55.0  & - \\
\bottomrule
\end{tabular}
}
\caption{Percentage of test set examples where the model generated the correct ground truth quotation for different folds of the test set. The \textsc{\faUser} and \textsc{\faBook} columns contain the accuracy of each model on the human-evaluated and close reading subsets of the data, respectively.} 
\label{tab:benchmark_accuracy}
\end{table}

The RELiC  dataset~\cite{thai-etal-2022-relic} consists of 78k excerpts of English-language literary analysis collected from scholarly journals where each excerpt includes a direct quotation from one of 79 primary source texts in the public domain. Each example in the dataset contains up to four sentences preceding the quotation and up to four sentences following the quotation. Together, these sentences of make up the ``context'' of the ground truth quotation. The primary source quotations may be up to five consecutive sentences in length.  

%Data about the books in the test and validation splits can be found in Table \td{ref}.
% , with the data for the train split books available in Appendix \ref{}.
% Because RELiC was designed as a retrieval task, the dataset also includes the full text of each primary source broken up into sentences.

\subsection{Adapting RELiC for long-context reasoning}
% The original data curation and processing steps for RELiC involved applying several programmatic heuristics, such as fuzzy-matching and regex, to the data to classify examples of literary analysis, identify the boundaries between literary analysis and ground truth quotations, and segment the data into sentences. 
While large, RELiC is also noisy, which necessitates several filtering steps before we can evaluate models on it. We implemented an extensive data preprocessing pipeline\footnote{All prompts, details of filtering heuristics, and descriptions of our manual validations are in Appendix A.} that involved several cleaning and filtering passes with \gptmini\ and \gpto\ supplemented by some programmatic heuristics, all geared towards mitigating the below issues:
\paragraph{Low data quality:} Some RELiC examples contain \textbf{OCR artifacts} present in the primary source texts that render the prefixes and suffixes ungrammatical. There are also some examples that are \textbf{misclassified} as literary analysis.
\paragraph{Model exploits:} Several aspects of RELiC examples can provide unintended cues to  models, allowing them to bypass the reasoning challenge. One was the \textbf{disclosure of the location} of the quote in the prefix or suffix. Another was \textbf{quote leakage}, or the appearance of part or all of the ground truth quotation in the context. Finally, \textbf{data contamination} was a concern--the literary analysis excerpts could appear in the training data of the LLMs we benchmarked and may have been memorized, or the ground truth quotation is so prevalent in the training data (because it belongs to a public domain novel) that the model is able to retrieve it without needing the primary source text at all.

\begin{figure*}[t]
  \centering
  \includegraphics[width=\textwidth]{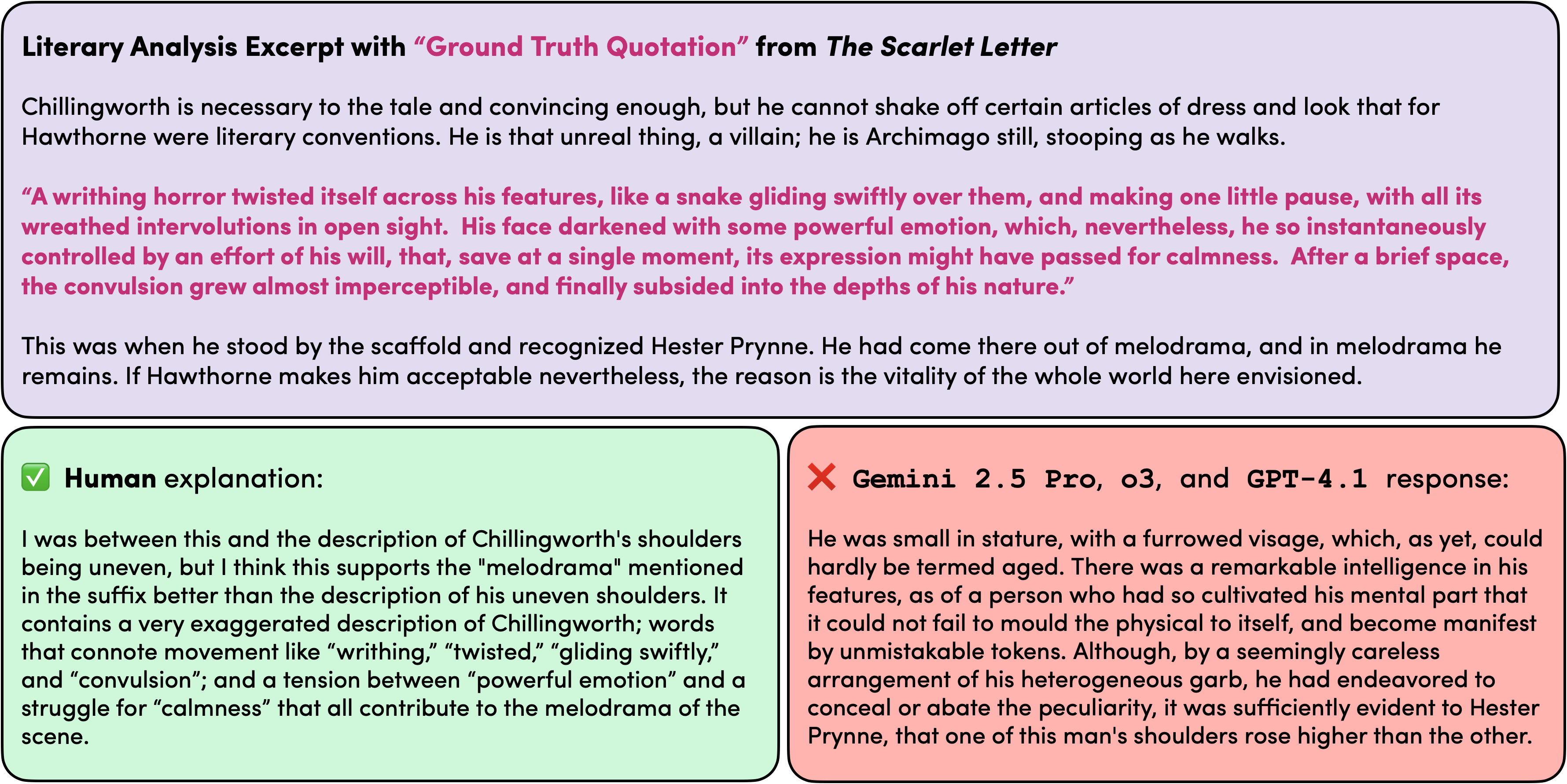}
  \caption{An example from long context RELiC where the \human\ annotator successfully retrieves the quotation while the top-performing LLMs fail.}
  \label{fig:llm_fail}
\end{figure*}

\paragraph{Human verification:} After filtering, we create a high-quality human-verified subset of the dataset by having one of the authors, who has a degree in English literature and has read all of the primary source novels manually review 400 filtered examples. This author marked 292 examples that were well-formed instances of literary analysis with a ground truth quotation that could be identified given only the book and literary analysis context. See Table \ref{tab:basic_summary_stats_dataset} for dataset statistics.

\paragraph{Dataset folds:} 
Our new dataset contains two labeled folds of special data, (1) the \faUser\ \textsc{Human Eval Set}: 40 examples attempted by our author, and (2) the \faBook\ \textsc{Close Reading Set}: 39 examples labeled by our author as examples of \textbf{close reading}, a literary analysis technique in which the reader considers ``linguistic elements, semantic aspects, syntax, rhetoric, structural elements, thematic, and generic references in the text''~\cite{close_reading}. A natural consequence of this interpretive technique is the repeated citation of parts of the ground truth quote in the context. Since these examples contain lexical overlap with the ground truth quotation, models can exploit this during evidence retrieval.

% \paragraph{} Finally, we randomly sampled 10,000 examples for the train set. We report statistics for all splits in Table \ref{} and for the test set folds in Table \ref{}. Though we do not utilize the train set in our experiments, we will release it to spur future research into increasing the performance of open source models on long-context reasoning tasks.

\section{Experimental Setup}
\label{sec:experiments}
We evaluated four closed-source and four open-weight models on the long-context RELiC dataset, prompting each in a zero-shot setting to retrieve the most fitting quotation for a literary analysis excerpt given the full primary source text. We tested two prompt types: (1) \textsc{Simple}, which requested only the quotation, and (2) \textsc{Explanation}, which first asked the model to justify its choice before selecting a quotation\footnote{Prompts \& inference details in Appendix \ref{sec:model-inf-details}.}. Our prompts are adapted from \citet{karpinska-etal-2024-nocha} in the \textsc{Nocha} long-context benchmark. Additionally, we implemented an embedding-based retrieval baseline using \gteqwen, the top-performing text embedding model on MTEB\footnote{\url{https://huggingface.co/spaces/mteb/leaderboard} accessed February 2025} at the time of writing.

\subsection{Human Evaluation}
The author who manually validated high quality examples also attempted 40 examples from four previously-read primary source novels. The author had access to a digital copy of the primary source text with a string-match search function and the corresponding Wikipedia summaries. In addition to selecting missing quotations, the author also wrote a short justification for each choice. The task took approximately 8 hours.

\subsection{Evaluation Scheme}
Our automatic evaluation of model responses applied partial ratio fuzzy matching (to account for minor typographical differences) to check that the model response was from the primary source text and to measure overlap between the response and the ground truth quotation\footnote{Further details can be found in Appendix \ref{sec:eval-details}.}. For the embedding baseline, we calculate recall@1.

% \FloatBarrier
\section{Results \& Analysis}
\label{sec:analysis}

\begin{figure*}[t]
  \centering
  \includegraphics[width=\textwidth]{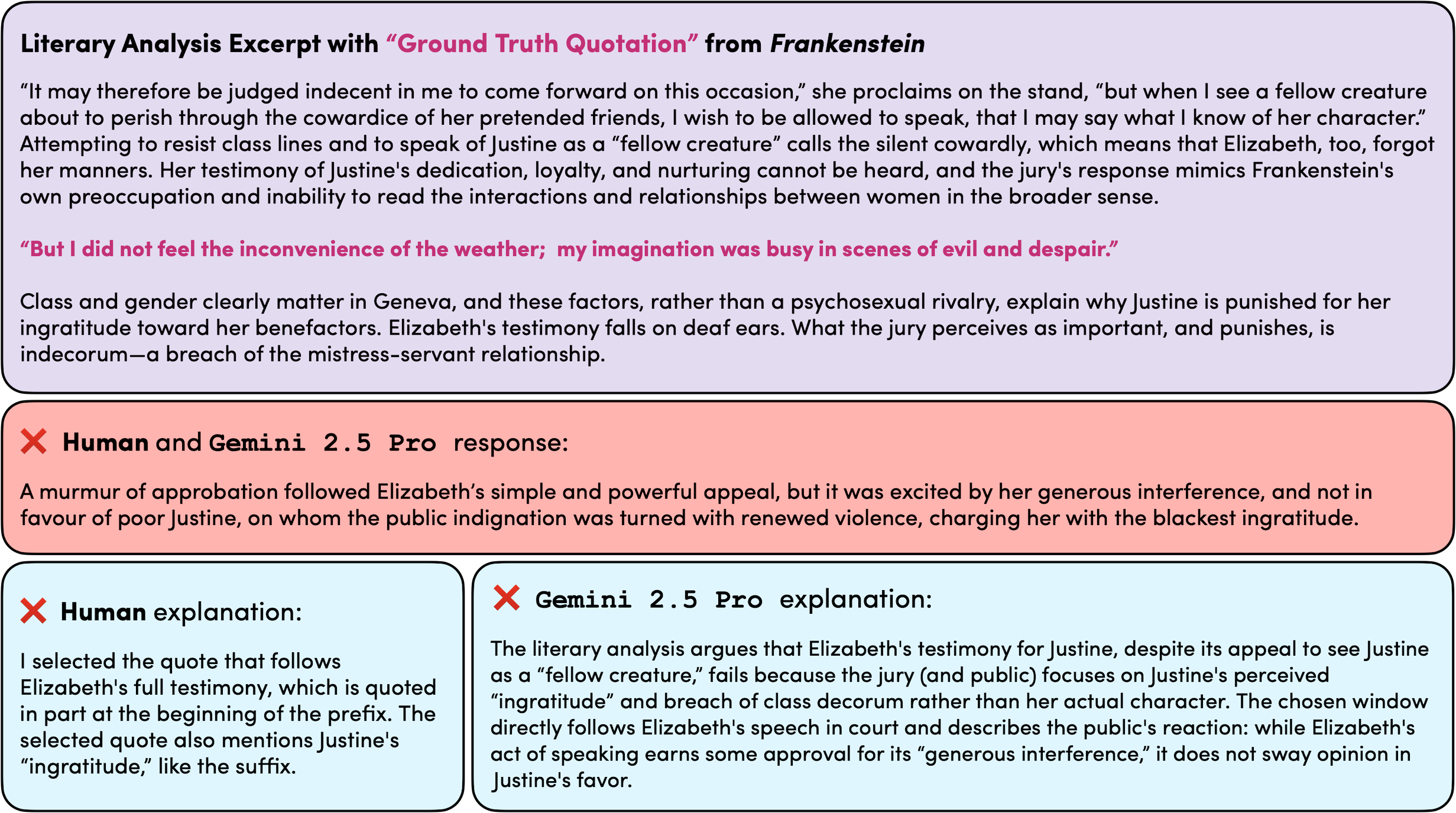}
  \caption{An example of long context RELiC where the \human\ annotator and \geminiprotwo\ both select the same quotation, but it is not the ground truth quotation.}
  \label{fig:alt_evidence}
\end{figure*}

Table \ref{tab:benchmark_accuracy} reports results for all evaluated models on the full set of 292 claims and the two folds. 

\paragraph{LLMs outperform \human\ and SOTA embedding baseline}
Google's \geminiprotwo\ reasoning model achieves better accuracy than our \human\ baseline in a fraction of the time---each API call took an average of 45 seconds vs. an average of 12 minutes for the \human\ expert. Both Google and OpenAI’s latest long-context models show substantial performance gains over their immediate predecessors on the literary evidence retrieval task. \geminiprotwo\ outperforms \geminiproone\ by a wide margin (64.7\% vs. 38.5\%), and OpenAI’s \gptfourone\ improves significantly over \gpto\ (51.0\% vs. 24.3\%). Similarly, \othree\ surpasses \oone\ (48.7\% vs. 32.2\%).
The embedding-based method achieved 4.5\% accuracy, only 1.6\% higher than the best recall@1 reported in the original RELiC paper from three years ago. LLMs’ success over embeddings suggests the importance of contextualized representations that can enable reasoning over entire primary source texts.

\paragraph{Closed-source outperform open-weight LLMs} Despite recent progress in open-weight LLM development, we observe a striking performance gap between closed-source and open-weight models on the literary evidence retrieval task. The best-performing open-weight model, \deepseek, achieves only 29.1\% accuracy---less than half the performance of the top closed-source model, \geminiprotwo\ (64.7\%).

\paragraph{Small LLMs can't capitalize on close reading examples} Nearly all models larger than 8B parameters show modest to significant improvements on the close reading fold, with some models improving by over 20\%. This effect is expected due to ground truth leakage in the the provided excerpt of literary analysis, but the 7B and 8B parameter open-weight models experience nearly no performance boost on this fold---the performance of \llamasmall\ actually suffers, suggesting that the smaller models lack the capacity to even exploit lexical overlap.

\paragraph{\textsc{Explanation} prompt provides a glimpse into model reasoning} We observed mixed results when querying non-reasoning LLMs with the \textsc{explanation} prompt. While \geminiproone\ improved with explanation-based reasoning, \gpto\ saw a decline. Recent reasoning LLMs such as \geminiprotwo\ and \othree\ use internal reasoning tokens that are not exposed via the API but influence the model's final output. These tokens are critical to the model's performance on complex tasks, but their inaccessibility complicates model evaluation and comparison. Applying the \textsc{explanation} prompt to these reasoning models preserved accuracy, despite the redundancy, while providing insight into the models' reasoning processes. See a comparison of human and model justifications in Figure \ref{fig:alt_evidence}.    

\paragraph{All models tend to overgenerate}
Despite prompts explicitly instructing models to select no more than five consecutive sentences from the primary source, we observe a consistent tendency across all evaluated models to produce significantly longer outputs. To quantify this behavior, we compute the length ratio, defined as the ratio of the model’s generated output length to the ground truth quotation length (measured in characters); values closer to 1.0 indicate greater adherence to the prompt constraints and minimal overgeneration. The length ratios are reported in Table \ref{tab:length_ratios}.

We find that all models overgenerate, including the human annotator (average ratio of 2.1), likely due to natural variance in interpreting sentence boundaries. However, language models consistently surpass this baseline: for instance, state-of-the-art models such as \geminiprotwo\ and \gptfourone\ have average ratios of 3.0 and 4.8, respectively. Smaller open models like \llamasmall\ and \llamalarge\ exhibit even more extreme overgeneration (ratios > 5.7), suggesting that weaker models may compensate for uncertainty by producing longer outputs. 

\paragraph{LLMs struggle with literary nuance}
In Figure \ref{fig:llm_fail}, we present a challenging example where all LLMs fail but our \human\ expert correctly identified the ground truth quotation from \textit{The Scarlet Letter}, a >80k-word novel. The context alludes to a description of the character Roger Chillingworth at a specific moment in the story (when he recognizes Hester Prynne). In their explanation, the \human\ expert mentions considering two different passages, but ultimately selecting the correct passage because it best demonstrates the ``melodrama" mentioned in the literary analysis. The expert highlights the features of the correct passage that demonstrate ``melodrama": an exaggerated character description, words that connote movement, and emotional tension. All three top-performing LLMs selected the other passage the \human\ expert was considering but ultimately eliminated, suggesting that even the best models still lack the expertise to navigate literary devices and more nuanced signals that inform expert literary interpretation. 

% \vspace{-10pt}
\paragraph{Models can identify alternative literary evidence} In Figure \ref{fig:alt_evidence}, we present an example where both \human\ and \geminiprotwo\ selected the same "incorrect" quote. The explanations reveal that both interpreted the literary analysis context as an introduction to a quote about the jury’s response. However, the ground truth quotation actually supports Frankenstein’s own preoccupation rather than the trial’s immediate outcome. This example highlights two key insights: (1) literary evidence retrieval is inherently interpretative, meaning that multiple passages may plausibly support a given claim, and (2) LLMs, like human readers, can surface alternative yet reasonable quotations that align with certain aspects of the analysis. While models may not always select the canonical answer, their ability to propose viable alternative evidence could be valuable for assisting literary scholars in exploring multiple textual connections.

\begin{table}[t!]
\small
\centering
\resizebox{0.48\textwidth}{!}{%
\begin{tabular}{lcc}
\toprule
\textsc{Model + Simple Prompt} & \textsc{\faUser} Accuracy \textsubscript{($n$=40)} & Avg. Length Ratio\\
\midrule
\geminiprotwo & 57.5 & 3.5\\
\othree & 47.5 & 3.5\\
\geminiproone & 22.5 & 2.8\\
\oone & 25.0 & 3.2\\
\gpto & 17.5 & 3.9\\
\qwenlarge & 7.5 & 3.2\\
\llamasmall  & 5.0 & 5.9\\
\qwensmall & 5.0 & 2.8\\
\llamalarge & 0.0 & 5.7\\
\midrule
\multicolumn{3}{l}{\textsc{Model + Explanation Prompt}} \\
\midrule
\geminiprotwo & \textbf{62.5} & 3.0\\
\gptfourone  & 47.5 & 4.8\\
\othree & 50.0 & 2.7\\
\geminiproone & 40.0 & 3.3\\
\claude & 32.5 & 4.0\\
\deepseek & 15.0 & 3.6\\
\gpto & 22.5 & 3.6\\
\qwenthreelarge & 20.0 & 2.7\\
\qwenthreesmall & 5.0 & 2.4\\
\othreemini & 10.0 & 3.5\\
\midrule
\multicolumn{3}{l}{\textsc{Baselines}} \\
\midrule
\human & 55.0 & \textbf{2.1}\\
\bottomrule
\end{tabular}
}
\caption{The average length ratios for each model, defined as the ratio of the length model generation to that of the ground truth (measured in characters). All models have an average ratio greater than that of the human annotator (2.1), indicating overgeneration.}
\label{tab:length_ratios}
\end{table}

\section{Related Work}
\label{sec:rel_work}

Our work builds on recent papers that apply and evaluate LLMs for computational literary analysis. Prior work has explored summarization or claim verification in novels~\cite{subbiah-etal-2024-storysumm,fables24,karpinska-etal-2024-nocha} or short stories~\cite{10.1162/tacl_a_00702}. Other more specific tasks, mainly in the short context setting, include  extracting narrative elements~\cite{shen-etal-2024-heart}, story arcs and turning points~\cite{tian-etal-2024-large-language}, character analysis~\cite{papoudakis-etal-2024-bookworm}, narrative discourse~\cite{piper-bagga-2024-using}, plot development~\cite{huot2024agents,xu-etal-2024-fine}, and creativity evaluation~\cite{10.1145/3613904.3642731}.

\section{Conclusion}
In this work, we introduced a long-context dataset derived from RELiC for evaluating literary evidence retrieval using large language models. We constructed a high-quality set of 292 examples, ensuring rigorous evaluation through filtering and human verification. Our results demonstrate that long-context LLMs outperform embedding-based retrieval methods and even a \human\ expert baseline. However, closer analysis reveals that model success is not always grounded in robust interpretative abilities: models tend to overgenerate and still struggle to grasp nuances in literature. But, they can also successfully identify additional evidence for literary claims. Finally, open-weight models lag far behind their closed-weight counterparts, suggesting that long-context capabilities alone are insufficient without strong interpretive reasoning. We release our data and evaluation code to spur research at the intersection of NLP and literary analysis.

\section*{Limitations}
As noted in the original paper, the RELiC dataset represents a limited, English-only subset of world literature, with a strong emphasis on the Western literary canon. To foster a more inclusive and representative benchmark, we aim to extend this work to cross-lingual literary analysis and retrieval on texts from historically underrepresented literary traditions. Expanding the dataset in this way would not only enhance the linguistic and cultural diversity of literary evidence retrieval but also improve the generalizability of models across different literary frameworks.

Additionally, our human evaluation was limited in scope and scale, as it was conducted by a single annotator, one of the authors of this paper. While this approach was practical given the intensive nature of literary evidence retrieval over full novels, it inherently reflects the knowledge, biases, and interpretive lens of a single individual. In the future, we aim to broaden our evaluation by incorporating multiple expert annotators, enabling a more comprehensive and diverse assessment of model performance on the task of complex literary reasoning.
\section*{Ethical Considerations}
While LLMs can assist in literary interpretation, they should not replace scholarly analysis conducted by trained experts. We acknowledge the potential for misuse of a system that can retrieve evidence for scholarly claims and emphasize that automated retrieval should be viewed as a complement to, not a substitute for, human literary scholarship.

As mentioned in the Limitations section, the primary source texts in our dataset are drawn from public domain works, which primarily reflect Western literary traditions. The results of this benchmark may not generalize to literary traditions outside the Anglophone canon, potentially reinforcing existing biases in computational literary studies.

\bibliography{custom}

\begin{thebibliography}{24}
\providecommand{\natexlab}[1]{#1}

\bibitem[{Anthropic(2025)}]{claude_sonnet_2025}
Anthropic. 2025.
\newblock \href {https://www.anthropic.com/news/claude-3-7-sonnet} {Claude 3.7 sonnet}.

\bibitem[{Chakrabarty et~al.(2024)Chakrabarty, Laban, Agarwal, Muresan, and Wu}]{10.1145/3613904.3642731}
Tuhin Chakrabarty, Philippe Laban, Divyansh Agarwal, Smaranda Muresan, and Chien-Sheng Wu. 2024.
\newblock \href {https://doi.org/10.1145/3613904.3642731} {Art or artifice? large language models and the false promise of creativity}.
\newblock In \emph{Proceedings of the 2024 CHI Conference on Human Factors in Computing Systems}, CHI '24, New York, NY, USA. Association for Computing Machinery.

\bibitem[{DeepSeek-AI et~al.(2025)DeepSeek-AI, Liu, Feng, Xue, Wang, Wu, Lu, Zhao, Deng, Zhang, Ruan, Dai, Guo, Yang, Chen, Ji, Li, Lin, Dai, Luo, Hao, Chen, Li, Zhang, Bao, Xu, Wang, Zhang, Ding, Xin, Gao, Li, Qu, Cai, Liang, Guo, Ni, Li, Wang, Chen, Chen, Yuan, Qiu, Li, Song, Dong, Hu, Gao, Guan, Huang, Yu, Wang, Zhang, Xu, Xia, Zhao, Wang, Zhang, Li, Wang, Zhang, Zhang, Tang, Li, Tian, Huang, Wang, Zhang, Wang, Zhu, Chen, Du, Chen, Jin, Ge, Zhang, Pan, Wang, Xu, Zhang, Chen, Li, Lu, Zhou, Chen, Wu, Ye, Ye, Ma, Wang, Zhou, Yu, Zhou, Pan, Wang, Yun, Pei, Sun, Xiao, Zeng, Zhao, An, Liu, Liang, Gao, Yu, Zhang, Li, Jin, Wang, Bi, Liu, Wang, Shen, Chen, Zhang, Chen, Nie, Sun, Wang, Cheng, Liu, Xie, Liu, Yu, Song, Shan, Zhou, Yang, Li, Su, Lin, Li, Wang, Wei, Zhu, Zhang, Xu, Xu, Huang, Li, Zhao, Sun, Li, Wang, Yu, Zheng, Zhang, Shi, Xiong, He, Tang, Piao, Wang, Tan, Ma, Liu, Guo, Wu, Ou, Zhu, Wang, Gong, Zou, He, Zha, Xiong, Ma, Yan, Luo, You, Liu, Zhou, Wu, Ren, Ren, Sha, Fu, Xu, Huang, Zhang, Xie, Zhang, Hao,
  Gou, Ma, Yan, Shao, Xu, Wu, Zhang, Li, Gu, Zhu, Liu, Li, Xie, Song, Gao, and Pan}]{deepseekai2025deepseekv3technicalreport}
DeepSeek-AI, Aixin Liu, Bei Feng, Bing Xue, Bingxuan Wang, Bochao Wu, Chengda Lu, Chenggang Zhao, Chengqi Deng, Chenyu Zhang, Chong Ruan, Damai Dai, Daya Guo, Dejian Yang, Deli Chen, Dongjie Ji, Erhang Li, Fangyun Lin, Fucong Dai, and 181 others. 2025.
\newblock \href {https://arxiv.org/abs/2412.19437} {Deepseek-v3 technical report}.
\newblock \emph{Preprint}, arXiv:2412.19437.

\bibitem[{{Gemini Team}(2024)}]{geminiteam2024gemini}
{Gemini Team}. 2024.
\newblock \href {https://arxiv.org/abs/2403.05530} {Gemini 1.5: Unlocking multimodal understanding across millions of tokens of context}.
\newblock \emph{Preprint}, arXiv:2403.05530.

\bibitem[{Google(2025)}]{google2025gemini25}
Google. 2025.
\newblock Gemini 2.5 pro.
\newblock Available at \url{https://deepmind.google/models/gemini/pro/}.

\bibitem[{Grattafiori et~al.(2024)Grattafiori, Dubey, Jauhri, Pandey, Kadian, Al-Dahle, Letman, Mathur, Schelten, Vaughan, Yang, Fan, Goyal, Hartshorn, Yang, Mitra, Sravankumar, Korenev, Hinsvark, Rao, Zhang, Rodriguez, Gregerson, Spataru, Roziere, Biron, Tang, Chern, Caucheteux, Nayak, Bi, Marra, McConnell, Keller, Touret, Wu, Wong, Ferrer, Nikolaidis, Allonsius, Song, Pintz, Livshits, Wyatt, Esiobu, Choudhary, Mahajan, Garcia-Olano, Perino, Hupkes, Lakomkin, AlBadawy, Lobanova, Dinan, Smith, Radenovic, Guzmán, Zhang, Synnaeve, Lee, Anderson, Thattai, Nail, Mialon, Pang, Cucurell, Nguyen, Korevaar, Xu, Touvron, Zarov, Ibarra, Kloumann, Misra, Evtimov, Zhang, Copet, Lee, Geffert, Vranes, Park, Mahadeokar, Shah, van~der Linde, Billock, Hong, Lee, Fu, Chi, Huang, Liu, Wang, Yu, Bitton, Spisak, Park, Rocca, Johnstun, Saxe, Jia, Alwala, Prasad, Upasani, Plawiak, Li, Heafield, Stone, El-Arini, Iyer, Malik, Chiu, Bhalla, Lakhotia, Rantala-Yeary, van~der Maaten, Chen, Tan, Jenkins, Martin, Madaan, Malo, Blecher,
  Landzaat, de~Oliveira, Muzzi, Pasupuleti, Singh, Paluri, Kardas, Tsimpoukelli, Oldham, Rita, Pavlova, Kambadur, Lewis, Si, Singh, Hassan, Goyal, Torabi, Bashlykov, Bogoychev, Chatterji, Zhang, Duchenne, Çelebi, Alrassy, Zhang, Li, Vasic, Weng, Bhargava, Dubal, Krishnan, Koura, Xu, He, Dong, Srinivasan, Ganapathy, Calderer, Cabral, Stojnic, Raileanu, Maheswari, Girdhar, Patel, Sauvestre, Polidoro, Sumbaly, Taylor, Silva, Hou, Wang, Hosseini, Chennabasappa, Singh, Bell, Kim, Edunov, Nie, Narang, Raparthy, Shen, Wan, Bhosale, Zhang, Vandenhende, Batra, Whitman, Sootla, Collot, Gururangan, Borodinsky, Herman, Fowler, Sheasha, Georgiou, Scialom, Speckbacher, Mihaylov, Xiao, Karn, Goswami, Gupta, Ramanathan, Kerkez, Gonguet, Do, Vogeti, Albiero, Petrovic, Chu, Xiong, Fu, Meers, Martinet, Wang, Wang, Tan, Xia, Xie, Jia, Wang, Goldschlag, Gaur, Babaei, Wen, Song, Zhang, Li, Mao, Coudert, Yan, Chen, Papakipos, Singh, Srivastava, Jain, Kelsey, Shajnfeld, Gangidi, Victoria, Goldstand, Menon, Sharma, Boesenberg,
  Baevski, Feinstein, Kallet, Sangani, Teo, Yunus, Lupu, Alvarado, Caples, Gu, Ho, Poulton, Ryan, Ramchandani, Dong, Franco, Goyal, Saraf, Chowdhury, Gabriel, Bharambe, Eisenman, Yazdan, James, Maurer, Leonhardi, Huang, Loyd, Paola, Paranjape, Liu, Wu, Ni, Hancock, Wasti, Spence, Stojkovic, Gamido, Montalvo, Parker, Burton, Mejia, Liu, Wang, Kim, Zhou, Hu, Chu, Cai, Tindal, Feichtenhofer, Gao, Civin, Beaty, Kreymer, Li, Adkins, Xu, Testuggine, David, Parikh, Liskovich, Foss, Wang, Le, Holland, Dowling, Jamil, Montgomery, Presani, Hahn, Wood, Le, Brinkman, Arcaute, Dunbar, Smothers, Sun, Kreuk, Tian, Kokkinos, Ozgenel, Caggioni, Kanayet, Seide, Florez, Schwarz, Badeer, Swee, Halpern, Herman, Sizov, Guangyi, Zhang, Lakshminarayanan, Inan, Shojanazeri, Zou, Wang, Zha, Habeeb, Rudolph, Suk, Aspegren, Goldman, Zhan, Damlaj, Molybog, Tufanov, Leontiadis, Veliche, Gat, Weissman, Geboski, Kohli, Lam, Asher, Gaya, Marcus, Tang, Chan, Zhen, Reizenstein, Teboul, Zhong, Jin, Yang, Cummings, Carvill, Shepard, McPhie,
  Torres, Ginsburg, Wang, Wu, U, Saxena, Khandelwal, Zand, Matosich, Veeraraghavan, Michelena, Li, Jagadeesh, Huang, Chawla, Huang, Chen, Garg, A, Silva, Bell, Zhang, Guo, Yu, Moshkovich, Wehrstedt, Khabsa, Avalani, Bhatt, Mankus, Hasson, Lennie, Reso, Groshev, Naumov, Lathi, Keneally, Liu, Seltzer, Valko, Restrepo, Patel, Vyatskov, Samvelyan, Clark, Macey, Wang, Hermoso, Metanat, Rastegari, Bansal, Santhanam, Parks, White, Bawa, Singhal, Egebo, Usunier, Mehta, Laptev, Dong, Cheng, Chernoguz, Hart, Salpekar, Kalinli, Kent, Parekh, Saab, Balaji, Rittner, Bontrager, Roux, Dollar, Zvyagina, Ratanchandani, Yuvraj, Liang, Alao, Rodriguez, Ayub, Murthy, Nayani, Mitra, Parthasarathy, Li, Hogan, Battey, Wang, Howes, Rinott, Mehta, Siby, Bondu, Datta, Chugh, Hunt, Dhillon, Sidorov, Pan, Mahajan, Verma, Yamamoto, Ramaswamy, Lindsay, Lindsay, Feng, Lin, Zha, Patil, Shankar, Zhang, Zhang, Wang, Agarwal, Sajuyigbe, Chintala, Max, Chen, Kehoe, Satterfield, Govindaprasad, Gupta, Deng, Cho, Virk, Subramanian, Choudhury,
  Goldman, Remez, Glaser, Best, Koehler, Robinson, Li, Zhang, Matthews, Chou, Shaked, Vontimitta, Ajayi, Montanez, Mohan, Kumar, Mangla, Ionescu, Poenaru, Mihailescu, Ivanov, Li, Wang, Jiang, Bouaziz, Constable, Tang, Wu, Wang, Wu, Gao, Kleinman, Chen, Hu, Jia, Qi, Li, Zhang, Zhang, Adi, Nam, Yu, Wang, Zhao, Hao, Qian, Li, He, Rait, DeVito, Rosnbrick, Wen, Yang, Zhao, and Ma}]{grattafiori2024llama3herdmodels}
Aaron Grattafiori, Abhimanyu Dubey, Abhinav Jauhri, Abhinav Pandey, Abhishek Kadian, Ahmad Al-Dahle, Aiesha Letman, Akhil Mathur, Alan Schelten, Alex Vaughan, Amy Yang, Angela Fan, Anirudh Goyal, Anthony Hartshorn, Aobo Yang, Archi Mitra, Archie Sravankumar, Artem Korenev, Arthur Hinsvark, and 542 others. 2024.
\newblock \href {https://arxiv.org/abs/2407.21783} {The llama 3 herd of models}.
\newblock \emph{Preprint}, arXiv:2407.21783.

\bibitem[{Huot et~al.(2024)Huot, Amplayo, Palomaki, Jakobovits, Clark, and Lapata}]{huot2024agents}
Fantine Huot, Reinald~Kim Amplayo, Jennimaria Palomaki, Alice~Shoshana Jakobovits, Elizabeth Clark, and Mirella Lapata. 2024.
\newblock Agents' room: Narrative generation through multi-step collaboration.
\newblock \emph{arXiv preprint arXiv:2410.02603}.

\bibitem[{Karpinska et~al.(2024)Karpinska, Thai, Lo, Goyal, and Iyyer}]{karpinska-etal-2024-nocha}
Marzena Karpinska, Katherine Thai, Kyle Lo, Tanya Goyal, and Mohit Iyyer. 2024.
\newblock \href {https://doi.org/10.18653/v1/2024.emnlp-main.948} {One thousand and one pairs: A {\textquotedblleft}novel{\textquotedblright} challenge for long-context language models}.
\newblock In \emph{Proceedings of the 2024 Conference on Empirical Methods in Natural Language Processing}, pages 17048--17085, Miami, Florida, USA. Association for Computational Linguistics.

\bibitem[{Kim et~al.(2024)Kim, Chang, Karpinska, Garimella, Manjunatha, Lo, Goyal, and Iyyer}]{fables24}
Yekyung Kim, Yapei Chang, Marzena Karpinska, Aparna Garimella, Varun Manjunatha, Kyle Lo, Tanya Goyal, and Mohit Iyyer. 2024.
\newblock Fables: Evaluating faithfulness and content selection in book-length summarization.
\newblock In \emph{Conference on Language Modeling}.

\bibitem[{Ohrvik(2024)}]{close_reading}
Ane Ohrvik. 2024.
\newblock \href {https://doi.org/10.1080/13642529.2024.2345001} {What is close reading? an exploration of a methodology}.
\newblock \emph{Rethinking History}, 28(2):238--260.

\bibitem[{OpenAI et~al.(2024)OpenAI, :, Hurst, Lerer, Goucher, Perelman, Ramesh, Clark, Ostrow, Welihinda, Hayes, Radford, Mądry, Baker-Whitcomb, Beutel, Borzunov, Carney, Chow, Kirillov, Nichol, Paino, Renzin, Passos, Kirillov, Christakis, Conneau, Kamali, Jabri, Moyer, Tam, Crookes, Tootoochian, Tootoonchian, Kumar, Vallone, Karpathy, Braunstein, Cann, Codispoti, Galu, Kondrich, Tulloch, Mishchenko, Baek, Jiang, Pelisse, Woodford, Gosalia, Dhar, Pantuliano, Nayak, Oliver, Zoph, Ghorbani, Leimberger, Rossen, Sokolowsky, Wang, Zweig, Hoover, Samic, McGrew, Spero, Giertler, Cheng, Lightcap, Walkin, Quinn, Guarraci, Hsu, Kellogg, Eastman, Lugaresi, Wainwright, Bassin, Hudson, Chu, Nelson, Li, Shern, Conger, Barette, Voss, Ding, Lu, Zhang, Beaumont, Hallacy, Koch, Gibson, Kim, Choi, McLeavey, Hesse, Fischer, Winter, Czarnecki, Jarvis, Wei, Koumouzelis, Sherburn, Kappler, Levin, Levy, Carr, Farhi, Mely, Robinson, Sasaki, Jin, Valladares, Tsipras, Li, Nguyen, Findlay, Oiwoh, Wong, Asdar, Proehl, Yang, Antonow,
  Kramer, Peterson, Sigler, Wallace, Brevdo, Mays, Khorasani, Such, Raso, Zhang, von Lohmann, Sulit, Goh, Oden, Salmon, Starace, Brockman, Salman, Bao, Hu, Wong, Wang, Schmidt, Whitney, Jun, Kirchner, de~Oliveira~Pinto, Ren, Chang, Chung, Kivlichan, O'Connell, O'Connell, Osband, Silber, Sohl, Okuyucu, Lan, Kostrikov, Sutskever, Kanitscheider, Gulrajani, Coxon, Menick, Pachocki, Aung, Betker, Crooks, Lennon, Kiros, Leike, Park, Kwon, Phang, Teplitz, Wei, Wolfe, Chen, Harris, Varavva, Lee, Shieh, Lin, Yu, Weng, Tang, Yu, Jang, Candela, Beutler, Landers, Parish, Heidecke, Schulman, Lachman, McKay, Uesato, Ward, Kim, Huizinga, Sitkin, Kraaijeveld, Gross, Kaplan, Snyder, Achiam, Jiao, Lee, Zhuang, Harriman, Fricke, Hayashi, Singhal, Shi, Karthik, Wood, Rimbach, Hsu, Nguyen, Gu-Lemberg, Button, Liu, Howe, Muthukumar, Luther, Ahmad, Kai, Itow, Workman, Pathak, Chen, Jing, Guy, Fedus, Zhou, Mamitsuka, Weng, McCallum, Held, Ouyang, Feuvrier, Zhang, Kondraciuk, Kaiser, Hewitt, Metz, Doshi, Aflak, Simens, Boyd,
  Thompson, Dukhan, Chen, Gray, Hudnall, Zhang, Aljubeh, Litwin, Zeng, Johnson, Shetty, Gupta, Shah, Yatbaz, Yang, Zhong, Glaese, Chen, Janner, Lampe, Petrov, Wu, Wang, Fradin, Pokrass, Castro, de~Castro, Pavlov, Brundage, Wang, Khan, Murati, Bavarian, Lin, Yesildal, Soto, Gimelshein, Cone, Staudacher, Summers, LaFontaine, Chowdhury, Ryder, Stathas, Turley, Tezak, Felix, Kudige, Keskar, Deutsch, Bundick, Puckett, Nachum, Okelola, Boiko, Murk, Jaffe, Watkins, Godement, Campbell-Moore, Chao, McMillan, Belov, Su, Bak, Bakkum, Deng, Dolan, Hoeschele, Welinder, Tillet, Pronin, Tillet, Dhariwal, Yuan, Dias, Lim, Arora, Troll, Lin, Lopes, Puri, Miyara, Leike, Gaubert, Zamani, Wang, Donnelly, Honsby, Smith, Sahai, Ramchandani, Huet, Carmichael, Zellers, Chen, Chen, Nigmatullin, Cheu, Jain, Altman, Schoenholz, Toizer, Miserendino, Agarwal, Culver, Ethersmith, Gray, Grove, Metzger, Hermani, Jain, Zhao, Wu, Jomoto, Wu, Shuaiqi, Xia, Phene, Papay, Narayanan, Coffey, Lee, Hall, Balaji, Broda, Stramer, Xu, Gogineni,
  Christianson, Sanders, Patwardhan, Cunninghman, Degry, Dimson, Raoux, Shadwell, Zheng, Underwood, Markov, Sherbakov, Rubin, Stasi, Kaftan, Heywood, Peterson, Walters, Eloundou, Qi, Moeller, Monaco, Kuo, Fomenko, Chang, Zheng, Zhou, Manassra, Sheu, Zaremba, Patil, Qian, Kim, Cheng, Zhang, He, Zhang, Jin, Dai, and Malkov}]{openai2024gpt4ocard}
OpenAI, :, Aaron Hurst, Adam Lerer, Adam~P. Goucher, Adam Perelman, Aditya Ramesh, Aidan Clark, AJ~Ostrow, Akila Welihinda, Alan Hayes, Alec Radford, Aleksander Mądry, Alex Baker-Whitcomb, Alex Beutel, Alex Borzunov, Alex Carney, Alex Chow, Alex Kirillov, and 401 others. 2024.
\newblock \href {https://arxiv.org/abs/2410.21276} {Gpt-4o system card}.
\newblock \emph{Preprint}, arXiv:2410.21276.

\bibitem[{OpenAI(2024)}]{openai2024o1}
OpenAI. 2024.
\newblock \href {https://platform.openai.com/docs/models} {O1 model card}.
\newblock Accessed: 2024-02-16.

\bibitem[{OpenAI(2025{\natexlab{a}})}]{gpt-4.1}
OpenAI. 2025{\natexlab{a}}.
\newblock \href {https://openai.com/index/gpt-4-1/} {Gpt-4.1}.

\bibitem[{OpenAI(2025{\natexlab{b}})}]{o3}
OpenAI. 2025{\natexlab{b}}.
\newblock \href {https://openai.com/index/introducing-o3-and-o4-mini/} {o3}.

\bibitem[{OpenAI(2025{\natexlab{c}})}]{openai2025o3mini}
OpenAI. 2025{\natexlab{c}}.
\newblock \href {https://cdn.openai.com/o3-mini-system-card-feb10.pdf} {Openai o3-mini system card}.
\newblock Accessed: 2025-02-15.

\bibitem[{Papoudakis et~al.(2024)Papoudakis, Lapata, and Keller}]{papoudakis-etal-2024-bookworm}
Argyrios Papoudakis, Mirella Lapata, and Frank Keller. 2024.
\newblock \href {https://doi.org/10.18653/v1/2024.findings-emnlp.258} {{B}ook{W}orm: A dataset for character description and analysis}.
\newblock In \emph{Findings of the Association for Computational Linguistics: EMNLP 2024}, pages 4471--4500, Miami, Florida, USA. Association for Computational Linguistics.

\bibitem[{Piper and Bagga(2024)}]{piper-bagga-2024-using}
Andrew Piper and Sunyam Bagga. 2024.
\newblock \href {https://doi.org/10.18653/v1/2024.wnu-1.4} {Using large language models for understanding narrative discourse}.
\newblock In \emph{Proceedings of the The 6th Workshop on Narrative Understanding}, pages 37--46, Miami, Florida, USA. Association for Computational Linguistics.

\bibitem[{Qwen et~al.(2025)Qwen, Yang, Yang, Zhang, Hui, Zheng, Yu, Li, Liu, Huang, Wei, Lin, Yang, Tu, Zhang, Yang, Yang, Zhou, Lin, Dang, Lu, Bao, Yang, Yu, Li, Xue, Zhang, Zhu, Men, Lin, Li, Tang, Xia, Ren, Ren, Fan, Su, Zhang, Wan, Liu, Cui, Zhang, and Qiu}]{qwen_qwen25_2025}
Qwen, An~Yang, Baosong Yang, Beichen Zhang, Binyuan Hui, Bo~Zheng, Bowen Yu, Chengyuan Li, Dayiheng Liu, Fei Huang, Haoran Wei, Huan Lin, Jian Yang, Jianhong Tu, Jianwei Zhang, Jianxin Yang, Jiaxi Yang, Jingren Zhou, Junyang Lin, and 24 others. 2025.
\newblock \href {https://doi.org/10.48550/arXiv.2412.15115} {Qwen2.5 {Technical} {Report}}.
\newblock \emph{arXiv preprint}.
\newblock ArXiv:2412.15115 [cs].

\bibitem[{Shen et~al.(2024)Shen, Mire, Park, Breazeal, and Sap}]{shen-etal-2024-heart}
Jocelyn Shen, Joel Mire, Hae~Won Park, Cynthia Breazeal, and Maarten Sap. 2024.
\newblock \href {https://doi.org/10.18653/v1/2024.emnlp-main.59} {{HEART}-felt narratives: Tracing empathy and narrative style in personal stories with {LLM}s}.
\newblock In \emph{Proceedings of the 2024 Conference on Empirical Methods in Natural Language Processing}, pages 1026--1046, Miami, Florida, USA. Association for Computational Linguistics.

\bibitem[{Subbiah et~al.(2024{\natexlab{a}})Subbiah, Ladhak, Mishra, Adams, Chilton, and McKeown}]{subbiah-etal-2024-storysumm}
Melanie Subbiah, Faisal Ladhak, Akankshya Mishra, Griffin~Thomas Adams, Lydia Chilton, and Kathleen McKeown. 2024{\natexlab{a}}.
\newblock \href {https://doi.org/10.18653/v1/2024.emnlp-main.557} {{STORYSUMM}: Evaluating faithfulness in story summarization}.
\newblock In \emph{Proceedings of the 2024 Conference on Empirical Methods in Natural Language Processing}, pages 9988--10005, Miami, Florida, USA. Association for Computational Linguistics.

\bibitem[{Subbiah et~al.(2024{\natexlab{b}})Subbiah, Zhang, Chilton, and McKeown}]{10.1162/tacl_a_00702}
Melanie Subbiah, Sean Zhang, Lydia~B. Chilton, and Kathleen McKeown. 2024{\natexlab{b}}.
\newblock \href {https://doi.org/10.1162/tacl_a_00702} {Reading subtext: Evaluating large language models on short story summarization with writers}.
\newblock \emph{Transactions of the Association for Computational Linguistics}, 12:1290--1310.

\bibitem[{Thai et~al.(2022)Thai, Chang, Krishna, and Iyyer}]{thai-etal-2022-relic}
Katherine Thai, Yapei Chang, Kalpesh Krishna, and Mohit Iyyer. 2022.
\newblock \href {https://doi.org/10.18653/v1/2022.acl-long.517} {{REL}i{C}: Retrieving evidence for literary claims}.
\newblock In \emph{Proceedings of the 60th Annual Meeting of the Association for Computational Linguistics (Volume 1: Long Papers)}, pages 7500--7518, Dublin, Ireland. Association for Computational Linguistics.

\bibitem[{Tian et~al.(2024)Tian, Huang, Liu, Jiang, Spangher, Chen, May, and Peng}]{tian-etal-2024-large-language}
Yufei Tian, Tenghao Huang, Miri Liu, Derek Jiang, Alexander Spangher, Muhao Chen, Jonathan May, and Nanyun Peng. 2024.
\newblock \href {https://doi.org/10.18653/v1/2024.emnlp-main.978} {Are large language models capable of generating human-level narratives?}
\newblock In \emph{Proceedings of the 2024 Conference on Empirical Methods in Natural Language Processing}, pages 17659--17681, Miami, Florida, USA. Association for Computational Linguistics.

\bibitem[{Xu et~al.(2024)Xu, Li, Yu, and Zhou}]{xu-etal-2024-fine}
Liyan Xu, Jiangnan Li, Mo~Yu, and Jie Zhou. 2024.
\newblock \href {https://doi.org/10.18653/v1/2024.acl-long.317} {Fine-grained modeling of narrative context: A coherence perspective via retrospective questions}.
\newblock In \emph{Proceedings of the 62nd Annual Meeting of the Association for Computational Linguistics (Volume 1: Long Papers)}, pages 5822--5838, Bangkok, Thailand. Association for Computational Linguistics.

\end{thebibliography}

\appendix
\section{Dataset Details}
\label{app:data_details}
We are using a publicly available dataset released by researchers who stated in their paper that they released their data to "facilitate further research in this direction." Statistics for the primary sources can be found in \ref{tab:book-details}.

\begin{table*}[t]
   \renewcommand{\arraystretch}{1.2} 
   \setlength{\tabcolsep}{6pt}
   \footnotesize
   \centering
   \begin{tabular}{l l c c c}
   \toprule
   \textbf{Book Title} & \textbf{Author} & \textbf{Publication Year} & \textbf{Token Count} & \textbf{Word Count} \\
   \midrule
   Brave New World            & Aldous Huxley       & 1932 & 91,472  & 65,278 \\
   What Maisie Knew           & Henry James         & 1897 & 124,544 & 95,988 \\
   Ethan Frome                & Edith Wharton       & 1911 & 45,038  & 34,926 \\
   Frankenstein               & Mary Shelley        & 1818 & 95,018  & 75,131 \\
   The Great Gatsby           & F. Scott Fitzgerald & 1925 & 63,683  & 48,972 \\
   The Awakening              & Kate Chopin         & 1899 & 66,574  & 49,932 \\
   The Scarlet Letter         & Nathaniel Hawthorne & 1850 & 112,355 & 83,311 \\
   \bottomrule
   \end{tabular}
   \caption{Books included in the dataset. The token count was provided as per \href{https://github.com/openai/tiktoken}{\texttt{tiktoken}} tokenization. }
   \label{tab:book-details}
\end{table*}

\section{LLM-Aided Data Processing}
\subsection{Filtering steps}
These cleaning and filtering steps were applied to the entire RELiC dataset in the following order:
\begin{enumerate}
    \item \verb|CLEAN| (\gptmini): We asked the model to remove any OCR artifacts and to ensure the context and ground truth quotation flow seamless and grammatically without changing any meaning. Additionally, we asked the model to remove any remaining in-line citations that revealed the page number of the ground truth quotation, which allowed us to preserve some examples that might have been caught by the \verb|LOCATION| filter below. See the prompt in Table \ref{tab:clean_filter}.
    \item \verb|LEAKAGE| (heuristic): If the sentences in the context preceding or following the ground truth quotation were fuzzy-matches (threshold: 95) with any text from the primary source, the example was excluded from our dataset.
    \item \verb|LIT ANALYSIS| (\gptmini): We asked the model to classify whether each RELiC instance was an example of literary analysis. If it was not, we excluded it from our dataset. See the prompt in Table \ref{tab:classify_lit_analysis}.
    \item \verb|LOCATION| (\gptmini): We asked the model if the context revealed the location of the ground truth quotation. If it did, we excluded it from our dataset. See the prompt in Table \ref{tab:quote_location_disclosure}.
    \item \verb|FIRST SENT| (heuristic): We identified cases where the ground truth quotation was the first sentence of its primary source novel. The intuition behind this filter (and the following two) was that the first sentence of a primary source might be famous or more likely to be quoted, and therefore easier to guess without needing to reason over the text or more likely to appear in training data.
    \item \verb|LAST SENT| (heuristic): Similarly, we identified cases where the ground truth quotation was the last sentence of its primary source novel.
    \item \verb|OUTLIER| (heuristic): We identified cases where the ground truth quotation was cited much more frequently than any of the other quotations from the same primary source novel.
    \item \verb|EZ2MEM| (\gpt): We asked the model to perform the RELiC task \textit{without} providing the primary source text and identified cases where the model was able to correctly generate at least one sentence of the ground truth quotation. 
\end{enumerate}
Note that we did not automatically exclude any of the examples identified by the last four filters. Though we did not use them in this project, we have left the labels in our dataset to facilitate future research.

\subsection{Manual validation of data}
We conducted multiple validations of the RELiC data for quality control using human annotations. 

\paragraph{Validation of LLM filters} To validate our LLM-aided approach to data preprocessing, we manually annotated 100 examples of RELiC and compared our judgments to keep or reject each example to the results of the LLM + heuristic filters. The f1-score of our filtering scheme was 89.8, with our scheme identifying 57 true positives, 30 true negatives, 6 false positives, and 7 false negatives.

\section{Model Inference Details}
\label{sec:model-inf-details}
\paragraph{API access details}
\begin{itemize}
    \item All OpenAI models were accessed via the OpenAI API (\url{https://platform.openai.com/}).
    \item  All Gemini models were accessed via the Google AI API (\url{https://ai.google.dev/}).
    \item \claude\ was accessed via the Vertex AI/Google Cloud (\url{https://cloud.google.com/vertex-ai?hl=en}).
    \item  DeepSeek was run via OpenRouter (\url{https://openrouter.ai/}).
\end{itemize}
The approximate total cost of running all closed-weight models for both development and evaluation was \$2k. 

\paragraph{Local inference details}
All Llama and Qwen models were run locally. The smaller open-weight models were run on 1 A100 GPU for a total of 14 hours, while the larger open-weight models were run on 4 A100 GPUs for a total of 16 hours. 

\paragraph{Generation hyperparameters}
The OpenAI reasoning models (\oone, \othreemini, and \othree) were a special case: the temperature parameter is fixed and the token generation limit includes the inaccessible reasoning tokens. As such, we set the token generation limit to 12,000 for \oone\ and \othreemini. For \othree, the token generation limit was set to 25,000 as recommended by OpenAI\footnote{\url{https://platform.openai.com/docs/guides/reasoning\#allocating-space-for-reasoning}}. For all three OpenAI reasoning models, we used the default \verb|medium| reasoning effort. 

For all other models, the token generation limit was set to 800 for \textsc{Simple} prompts and 1,200 for \textsc{Explanation} prompts. Temperature was set to 0.0 for all models except for \qwenthreelarge\ and \qwenthreesmall, where temperature was set to the default 0.6.

For a list of the models and their checkpoints, see Table \ref{tab:models}.

\begin{table}[t!]
\centering
\small
\resizebox{0.48\textwidth}{!}{%
\begin{tabular}{l c l c  c}
\toprule
\textsc{Model} & \textsc{Context} & \textsc{Avail.} & \textsc{Checkpoints}  \\
\midrule
\href{https://platform.openai.com/docs/models}{\oone} & 200k & \faLock & \texttt{o1-2024-12-17}  \\
\href{https://platform.openai.com/docs/models}{\othree} & 200k &  \faLock & \texttt{o3-2025-04-16} \\
\href{https://platform.openai.com/docs/models}{\othreemini} & 200k & \faLock & \texttt{o3-mini-2025-01-31} \\
\href{https://platform.openai.com/docs/models}{\gpto} & 128k & \faLock & \texttt{gpt-4o-2024-11-20} \\
\href{https://platform.openai.com/docs/models}{\gptfourone} & 1M & \faLock & \texttt{gpt-4.1-2025-01-31} \\
\href{https://cloud.google.com/vertex-ai/generative-ai/docs/model-reference/gemini}{\geminiproone} & 1M & \faLock & \texttt{gemini-1.5-pro-002} \\
\href{https://cloud.google.com/vertex-ai/generative-ai/docs/model-reference/gemini}{\geminiprotwo} & 1M & \faLock & \texttt{gemini-2.5-pro-preview-05-06} \\
\href{https://docs.anthropic.com/en/docs/about-claude/models/overview}{\claude} & 200k & \faLock & \texttt{claude-3-7-sonnet@20250219} \\
\href{https://huggingface.co/meta-llama/Llama-3.3-70B-Instruct}{\llamalarge} & 128k & \faUnlock & \texttt{Llama-3.3-70B-Instruct} \\
\href{https://huggingface.co/meta-llama/Llama-3.1-8B-Instruct}{\llamasmall} & 128k & \faUnlock & \texttt{
Llama-3.1-8B-Instruct }  \\
\href{https://huggingface.co/Qwen/Qwen2.5-72B-Instruct}{\qwenlarge} & 128k & \faUnlock & \texttt{Qwen2.5-72B-Instruct} \\
\href{https://huggingface.co/Qwen/Qwen2.5-7B-Instruct}{\qwensmall} & 128k & \faUnlock & \texttt{Qwen2.5-7B-Instruct} \\
\href{https://huggingface.co/Qwen/Qwen3-8B}{\qwenthreesmall} & 131k* & \faUnlock & \texttt{Qwen3-8B} \\
\href{https://huggingface.co/Qwen/Qwen3-32B}{\qwenthreelarge} & 131k* & \faUnlock & \texttt{Qwen3-32B} \\
\midrule
\href{https://huggingface.co/Alibaba-NLP/gte-Qwen2-7B-instruct}{\gteqwenparam} & 32k & \faUnlock & \texttt{gte-Qwen2-7B-instruct}  \\
\bottomrule
\end{tabular}
}
\caption{The upper rows display the evaluated LLMs, while the bottom row displays the text embedding model used for the baseline. Context lengths marked with * were extended with YaRN.}
\label{tab:models}
\end{table}

\section{Evaluation Scheme Details}
\label{sec:eval-details}
We use the \verb|rapidfuzz| Python package for our fuzzy match evaluation with a threshold of 95 for checking the existence of the model response in the primary source and a threshold of 90 for checking the overlap between the model response and the ground truth quotation. The fuzzy matching allows for small typographical differences between the primary source and the model outputs (e.g. different types of quotation marks). Thresholds were determined after manual inspection of outputs at varying thresholds.

% \section{Additional Example for \textsc{Human} vs. \textsc{Model}}
% \label{sec:hum_vs_model_extra}

\begin{table*}[t]
   \renewcommand{\arraystretch}{1.2} 
   \setlength{\tabcolsep}{4pt}
   \footnotesize
   \centering
   \begin{tabular}{c p{15cm}}
   \toprule
        & \multicolumn{1}{c}{\bf Prompt (Simple)} \\
   \midrule
    \noalign{\vskip 1mm}
    & \texttt{You are provided with the full text of \textbf{[book\_title]} and an excerpt of literary analysis that directly cites \textbf{[book\_title]} with the cited quotation represented as \textless MASK\textgreater.} \\[3mm]
    & \texttt{Your task is to carefully read the text of \textbf{[book\_title]} and the excerpt of literary analysis, then select a window from \textbf{[book\_title]} that most appropriately replaces \textless MASK\textgreater as the cited quotation by providing textual evidence for any claims in the literary analysis.} \\[1mm]
    & \texttt{The excerpt of literary analysis should form a valid argument when \textless MASK\textgreater is replaced by the window from \textbf{[book\_title]}.} \\[1mm]
    & \texttt{\textless full\_text\_of\_[book\_title\_snake\_case]\textgreater \textbf{[book\_sentences]} \textless /full\_text\_of\_[book\_title\_snake\_case]\textgreater} \\[1mm]
    & \texttt{\textless literary\_analysis\_excerpt\textgreater \textbf{[lit\_analysis\_excerpt]} \textless /literary\_analysis\_excerpt\textgreater} \\[1mm]
    & \texttt{Identify the window that best supports the claims being made in the excerpt of literary analysis. The window should contain no more than 5 consecutive sentences from \textbf{[book\_title]}.} \\[1mm]
    & \texttt{Provide your final answer in the following format:} \\[1mm]
    & \texttt{\textless window\textgreater \textbf{YOUR SELECTED WINDOW} \textless /window\textgreater} \\[1mm]
   \bottomrule
   \end{tabular}
\caption{Prompt template for literary evidence retrieval (Simple).}
\label{tab:inference_simple}
\end{table*}

\begin{table*}[t]
   \renewcommand{\arraystretch}{1.2} 
   \setlength{\tabcolsep}{4pt}
   \footnotesize
   \centering
   \begin{tabular}{c p{15cm}}
   \toprule
        & \multicolumn{1}{c}{\bf Prompt w/ Explanations} \\
   \midrule
    \noalign{\vskip 1mm}
    & \texttt{You are provided with the full text of \textbf{[book\_title]} and an excerpt of literary analysis that directly cites \textbf{[book\_title]} with the cited quotation represented as \textless MASK\textgreater.} \\[3mm]
    & \texttt{Your task is to carefully read the text of \textbf{[book\_title]} and the excerpt of literary analysis, then select a window from \textbf{[book\_title]} that most appropriately replaces \textless MASK\textgreater as the cited quotation by providing textual evidence for any claims in the literary analysis.} \\[1mm]
    & \texttt{The excerpt of literary analysis should form a valid argument when \textless MASK\textgreater is replaced by the window from \textbf{[book\_title]}.} \\[1mm]
    & \texttt{\textless full\_text\_of\_[book\_title\_snake\_case]\textgreater \textbf{[book\_sentences]} \textless /full\_text\_of\_[book\_title\_snake\_case]\textgreater} \\[1mm]
    & \texttt{\textless literary\_analysis\_excerpt\textgreater \textbf{[lit\_analysis\_excerpt]} \textless /literary\_analysis\_excerpt\textgreater} \\[1mm]
    & \texttt{First, provide an explanation of your decision marking process in no more than one paragraph.} \\[1mm]
    & \texttt{Then, identify the window that best supports the claims being made in the excerpt of literary analysis. The window should contain no more than 5 consecutive sentences from \textbf{[book\_title]}.} \\[1mm]
    & \texttt{Provide your final answer in the following format:} \\[1mm]
    & \texttt{\textless explanation\textgreater \textbf{YOUR EXPLANATION} \textless /explanation\textgreater} \\[1mm]
    & \texttt{\textless window\textgreater \textbf{YOUR SELECTED WINDOW} \textless /window\textgreater} \\[1mm]
   \bottomrule
   \end{tabular}
\caption{Prompt template for literary evidence retrieval with explanation.}
\label{tab:inference_explain}
\end{table*}

\begin{table*}[t]
   \renewcommand{\arraystretch}{1.2} 
   \setlength{\tabcolsep}{4pt}
   \footnotesize
   \centering
   \begin{tabular}{c p{15cm}}
   \toprule
        & \multicolumn{1}{c}{\bf Prompt for Cleaning RELiC Examples} \\
   \midrule
    \noalign{\vskip 1mm}
    & \texttt{I want to create a dataset for Natural Language Processing that consists of excerpts of literary analysis that quote directly from a primary source text.} \\[3mm]
    & \texttt{I have already collected windows of literary analysis that quote from many different primary sources in the public domain. The part of the excerpt before the quotation is called the "prefix," and the part of the excerpt after the quotation is called the "suffix." Because I collected the data from PDFs with OCR, the prefixes and suffixes contain artifacts like in-line citations, page numbers, chapter names, headers, and footers that I need to remove.} \\[3mm]
    & \texttt{Here is an example:} \\[1mm]
    & \texttt{\textless prefix\textgreater \textbf{[prefix]} \textless /prefix\textgreater} \\[1mm]
    & \texttt{\textless ground\_truth\_quotation\textgreater \textbf{[ground\_truth\_quotation]} \textless /ground\_truth\_quotation\textgreater} \\[1mm]
    & \texttt{\textless suffix\textgreater \textbf{[suffix]} \textless /suffix\textgreater} \\[1mm]
    & \texttt{Please follow the following guidelines to help me clean and filter this example:} \\[1mm]
    & \texttt{1. Remove all textual artifacts from the OCR process by deleting them. Artifacts include page numbers, chapter headings, footnotes, image captions, etc.} \\[1mm]
    & \texttt{2. Correct the grammar, punctuation, and spelling of the prefix and suffix without altering the meaning so that the primary source quotation fits seamlessly and grammatically between the prefix and suffix.} \\[1mm]
    & \texttt{3. Remove all in-line citations following quotations, especially those that say things like "(my emphasis)."} \\[1mm]
    & \texttt{Respond with ONLY the cleaned prefix and suffix in the following format:} \\[1mm]
    & \texttt{\textless clean\_prefix\textgreater \textbf{CLEAN PREFIX} \textless /clean\_prefix\textgreater} \\[1mm]
    & \texttt{\textless clean\_suffix\textgreater \textbf{CLEAN SUFFIX} \textless /clean\_suffix\textgreater} \\[1mm]
   \bottomrule
   \end{tabular}
\caption{Prompt template for cleaning RELiC examples}
\label{tab:clean_filter}
\end{table*}

\begin{table*}[t]
   \renewcommand{\arraystretch}{1.2} 
   \setlength{\tabcolsep}{4pt}
   \footnotesize
   \centering
   \begin{tabular}{c p{15cm}}
   \toprule
        & \multicolumn{1}{c}{\bf Prompt for Classifying Literary Analysis} \\
   \midrule
    \noalign{\vskip 1mm}
    & \texttt{I want to create a dataset for Natural Language Processing that consists of excerpts of literary analysis that quote directly from a primary source text.} \\[3mm]
    & \texttt{Other texts can contain quotes from primary sources, like biographies of authors. However, I only want examples of literary analysis in the dataset.} \\[3mm]
    & \texttt{Please determine whether the following excerpt is an example of literary analysis. If it is literary analysis, respond with TRUE. Otherwise, respond with FALSE.} \\[3mm]
    & \texttt{Here is the excerpt:} \\[1mm]
    & \texttt{\textless excerpt\textgreater \textbf{[clean\_prefix]} \textbf{[answer\_quote]} \textbf{[clean\_suffix]} \textless /excerpt\textgreater} \\[1mm]
    & \texttt{Respond with ONLY your answer in the following format:} \\[1mm]
    & \texttt{\textless answer\textgreater \textbf{YOUR ANSWER} \textless /answer\textgreater} \\[1mm]
   \bottomrule
   \end{tabular}
\caption{Prompt template for classifying literary analysis}
\label{tab:classify_lit_analysis}
\end{table*}

\begin{table*}[t]
   \renewcommand{\arraystretch}{1.2} 
   \setlength{\tabcolsep}{4pt}
   \footnotesize
   \centering
   \begin{tabular}{c p{15cm}}
   \toprule
        & \multicolumn{1}{c}{\bf Prompt for Identifying Quote Location Disclosure} \\
   \midrule
    \noalign{\vskip 1mm}
    & \texttt{I want to create a dataset for a Natural Language Processing task called "Literary Evidence Retrieval."} \\[3mm]
    & \texttt{The input has two parts:} \\[1mm]
    & \texttt{(1) An excerpt of literary analysis that quotes 1 to 5 full sentences from a primary source text, but the quote is replaced with "[MASK]".} \\[1mm]
    & \texttt{(2) The full text of the primary source.} \\[3mm]
    & \texttt{The output should be the correct quotation of 1 to 5 full sentences from the primary source—this is the ground truth.} \\[3mm]
    & \texttt{I have already collected windows of literary analysis that quote from many different primary sources in the public domain.} \\[3mm]
    & \texttt{The part of the excerpt before the quotation is called the "prefix," and the part after is called the "suffix."} \\[3mm]
    & \texttt{To avoid giving the model hints, it's extremely important that the prefix and suffix do not mention the chapter where the ground truth quotation is located.} \\[3mm]
    & \texttt{Please determine whether the following instance contains the chapter location of the ground truth quotation in either the prefix or the suffix. If it does, respond with TRUE; otherwise, respond with FALSE.} \\[3mm]
    & \texttt{The prefix and suffix may contain the location of other quotations in the passage. I only want to identify if the location of the ground truth quotation is revealed.} \\[3mm]
    & \texttt{Look out for phrases like "In the prologue" or "In Chapter..."} \\[3mm]
    & \texttt{Here is the instance:} \\[1mm]
    & \texttt{\textless prefix\textgreater \textbf{[prefix]} \textless /prefix\textgreater} \\[1mm]
    & \texttt{\textless ground\_truth\_quotation\textgreater \textbf{[ground\_truth\_quotation]} \textless /ground\_truth\_quotation\textgreater} \\[1mm]
    & \texttt{\textless suffix\textgreater \textbf{[suffix]} \textless /suffix\textgreater} \\[1mm]
    & \texttt{Respond with ONLY your answer in the following format:} \\[1mm]
    & \texttt{\textless answer\textgreater \textbf{YOUR ANSWER} \textless /answer\textgreater} \\[1mm]
   \bottomrule
   \end{tabular}
\caption{Prompt template for identifying quote location disclosure}
\label{tab:quote_location_disclosure}
\end{table*}

\section{Use of AI Assistants}
The authors used Github Copilot for coding assistance during their experiments and ChatGPT for assistance in formatting \LaTeX\ in the writing of this paper.

\end{document}